\documentclass[runningheads]{llncs}
\usepackage{graphicx}
\begin{document}
\title{Machine Learning to Estimate Gross Loss of Jewelry
for Wax Patterns}
%
%
\author{Mihir Jain\inst{1}\and
	Kashish Jain\inst{2} \and
	Sandip Mane\inst{3}}
\authorrunning{M. Jain et al.}
%
\institute{Purdue School of Industrial Engineering, West Lafayette, IN 47906, USA \email{jain574@purdue.edu} \and
  Sardar Patel Institute of Technology, Mumbai, 400058, India \email{kashish.jain@spit.ac.in}
  \and
  D.J Sanghvi College of Engineering, Mumbai, 400056, India \email{sandip.mane@djsce.ac.in}\\}

	\maketitle              
	\begin{abstract}
		In mass manufacturing of jewellery, the gross loss is estimated before manufacturing to calculate the wax weight of the pattern that would be investment casted to make multiple identical pieces of jewellery. Machine learning is a technology that is a part of AI which helps create a model with decision-making capabilities based on a large set of user-defined data. In this paper, the authors found a way to use Machine Learning in the jewellery industry to estimate this crucial Gross Loss. Choosing a small data set of manufactured rings and via regression analysis, it was found out that there is a potential of reducing the error in estimation from ±2-3 to ±0.5- using ML Algorithms from historic data and attributes collected from the CAD file during the design phase itself. To evaluate the approach's viability, additional study must be undertaken with a larger data set.
				
		\keywords{CAD  \and Gross Loss \and Jewelry \and Machine Learning \and Wax Model.}
	\end{abstract}
	\section{Introduction}
	Loss is an inevitable component of manufacturing. In the manufacturing of jewellery from precious metals, accounting and calculating the losses is a very crucial. Gross Loss of jewellery is the total metal loss during its manufacturing. Loss in metal happens during casting, filing, polishing, setting and at almost every stage. Even though most of this lost metal is recovered and refined in the refinery to get a recovery of 92\%, on average, these losses are extremely crucial not to be accounted for.
		
	The loss on each piece of jewellery varies, based on various factors. Estimating this gross loss beforehand was very crucial for the manufacturing of that jewellery. This estimated gross loss was used for while pulling wax patterns during the process of injection moulding \cite{casting}. Jewelry made from the heavier wax piece will have surplus metal that must be filed down and recovered later, which is a waste of time and materials because only some of the metal will be recovered. Therefore, estimating the total loss provides a general estimate of the wax weight and can be used as a guide for how each procedure should be carried out.
		
	In a production process, a step wise loss of each of step of manufacturing is collected. This is done by weighing the jewelry after each step. Hence after the jewelry has been manufactured it can assess the final data of gross loss that the company bore. Total recovery that was done was also considered, and added to the database.
		
	This gross loss found out was further collected out of which a wide set of databases is manufactured by an in-house engineer. Calculations based on current trends are made where a few other variables are also taken into consideration.
		
	Variables like, weight of the final product, metal type (White Gold, Yellow Gold, Pink Gold, Silver, Platinum and Palladium), cartage of metal (8k, 9k, 10k, 12k, 14k, 18k, 20k etc.), the customer for whom the jewelry is being manufactured, the setting of diamond (whether the piece is handset or wax set) and of course the type of jewelry it is (whether it is a ring, a pendant, an earring, a bracelet or a bangle.)
		
	Currently, the estimation comes with a variance of ±4-5\%. Hence there is a scope here by which, using the powerful tools of Machine Learning \cite{ml1,ml2,ml3,ml4} we can consider the variable constants to find out the gross loss in jewelry. These variable constants can most often than not be fetched directly from the CAD files which are made way before the actual manufacturing process even begins.
		
	The aim of the paper is to estimate the gross loss of jewelry at the CAD level with greater and repeatable accuracy using machine learning algorithms. This paper will systematically narrow down the variables responsible for gross loss of jewelry during its manufacturing, create a machine learning model that predicts the final gross loss based on the data collected from the CAD file generated before manufacturing and ensure greater accuracy of the model as compared to the traditional methods of estimating loss.
		
	\section{Methodology}
    As the project is a proof of concept, it only takes into account 26 rings as a sample size. This project will only use information from the last several months of production for all ring kinds for which CAD files were available (developed in Rhino 3D \cite{rhino}) and for which the company knew the associated gross losses. It is important to highlight that only information that could be shown publicly has been included in this report. There were notably three stages to the project's execution.
		
	\subsection{Creating the Dataset}
	The first phase comprised of selecting all possible attribute of the rings from the CAD file and listing them down with their corresponding values on an excel sheet. This data was paired with its corresponding historic gross loss.
	\begin{table}[!ht]
		\centering
		\caption{Parameters of the Dataset}
		\begin{tabular}{|l|l|l|} 
			\hline
			\textbf{\#} & \textbf{Attribute}              & \textbf{Datatype}     \\ 
			\hline
			1           & Volume                          & mm\textsuperscript{3} \\ 
			2           & Surface Area                    & mm\textsuperscript{2} \\ 
			3           & Metal                           & Karat-metal           \\ 
			4               & Weight/ Piece
			(Estimated)   & gm                            \\ 
			5           & Total Lot Quantity              & integer               \\ 
			6               & Total Weight
			of Lot         & gm                            \\ 
			7           & Inner Diameter                  & mm                    \\ 
			8           & Outer Diameter                  & mm                    \\ 
			9               & Minimum Shank
			Thickness     & mm                            \\ 
			10              & Maximum Shank
			Thickness     & mm                            \\ 
			11          & Minimum Shank Width             & mm                    \\ 
			12          & Maximum Shank Width             & mm                    \\ 
			13          & Total Height                    & mm                    \\ 
			14          & Top Height                      & mm                    \\ 
			15          & Number of Components            & integer               \\ 
			16          & Number of Rings                 & integer               \\ 
			17          & Tone                            & 1/2/3                 \\ 
			18          & True Miracle                    & binary                \\ 
			19          & No. of True Miracle             & integer               \\ 
			20          & Diamond – Handset and Wax Set & integer               \\ 
			21          & Filigree                        & binary                \\ 
			22          & J Back                          & binary                \\ 
			23          & Gallery                         & binary                \\ 
			24          & Fake Beads                      & integer               \\ 
			25          & Plating                         & binary                \\
			\hline
		\end{tabular}
	\end{table}
		
	\subsection{Preparation of Data}
	The compiled data was obtained from the CAD files. This data had irrelevant parameters that are currently unknown but will be filtered out through processing. The reason why all possible data was collected was to avoid any human generated discrepancies in the very first stage of the project. Even though 26 is a small number for a machine learning algorithm, its corresponding volume would still suffice to give us the proof of concept required to carry on with the project. But in an ideal situation, the number of rows should be 4x the number of columns. So, it can safely consider that the results obtained will only improve when the amount of data of rings increases. Before testing algorithms, the data was checked for any missing values and such values were designated a weighted average value. Feature scaling was done using Standard Scaling which standardize features by removing the mean and scaling to unit variance \cite{scikit-learn}. The standard score of a sample x is calculated as:
	\begin{equation}
		z = (x - u) / s
	\end{equation}
	where u is the mean of the training samples, and s is the standard deviation of the training samples.

	\subsection{Models}
	The above-mentioned data was split tested, and the mean value graph of each feature was derived. The data set was split as 80-20, where 80 was the train dataset and 20 was the test dataset. After the data set is split, feature scaling was done to make the comparison easier. This data was the processed through various algorithms.
	\begin{enumerate}
		\item \textit{Linear Regression Algorithm}
		      \begin{figure}[h]
		      	\centering
		      	\includegraphics[width=0.3\paperwidth]{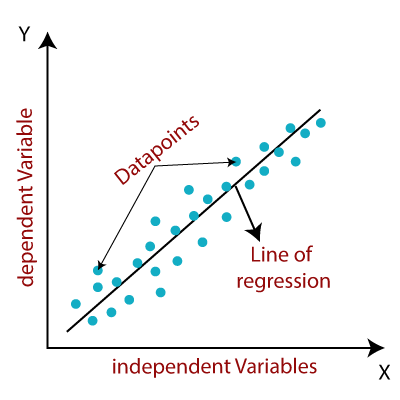}
		      	\caption{Architecture of Linear Regression Algorithm}
		      \end{figure}
		      \begin{enumerate}
		      	\item Linear Regression \cite{lr_origin} is a supervised learning-based machine learning technique. One of its functions is to carry out a regression analysis. Through the use of independent variables, the regression model can predict a desired outcome. Its primary function is to investigate causal links between factors and predicting. 
		      	\item Predicting a value (y) of a dependent variable (x) from known values (y) of independent variables (x) is the job of linear regression (x). Accordingly, this method of regression establishes a linear connection between the input variable (x) and the output variable (y) (output). Linear Regression perfectly describes this method \cite{lr}. Hypothesis function for Linear Regression :
		      	      \begin{equation}
		      	      	y = \theta_1 + \theta_2.x
		      	      \end{equation}
		      	\item When training the model – it fits the best line to predict the value of y for a given value of x. The model gets the best regression fit line by finding the best $\theta$\textsubscript{1} and $\theta$\textsubscript{2} values where $\theta$\textsubscript{1} is the intercept and $\theta$\textsubscript{2} is the coefficient of x. 
		      	      		      	      	      
		      	\item The best fit line is obtained by locating the optimal values of $\theta$\textsubscript{1} and $\theta$\textsubscript{2}. When our model is used for prediction, it will give us y as a function of x.
		      \end{enumerate}
		      		      
		\item \textit{Random Forest Regression}
		      		        
		      \begin{figure}[h]
		      	\centering
		      	\includegraphics[width=0.4\paperwidth]{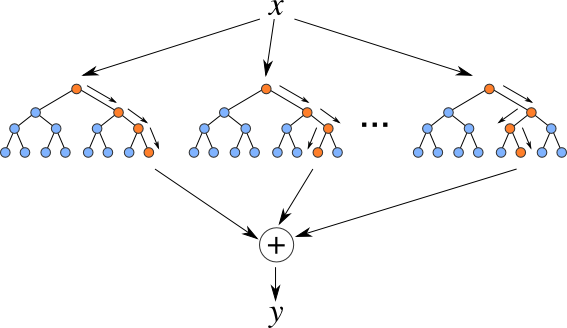}
		      	\caption{Architecture of Random Forest Algorithm}
		      \end{figure}
		      \begin{enumerate}
		      	\item Random Forest Regression \cite{rf} algorithm is an example of a supervised learning algorithm that use the ensemble learning approach of regression. By combining the results of several different machine learning algorithms, an ensemble learning method can produce a more precise forecast than any one of them could on its own.
		      	\item Random Forest relies on the "wisdom of the crowds" principle, which states that a large number of independent models working in concert can achieve better results than any of their parts working alone.
		      	\item This is owing to the fact that the trees buffer one another from their particular errors. Since a random forest is completely random, there is no communication between the trees that make up the forest. Random forest is an estimator technique that takes the outputs of multiple decision trees, compiles them, and then generates the ideal answer for the given situation. \cite{rf_website}
		      \end{enumerate}
		      		      
		\item \textit{Decision Tree Regression}
		      \begin{figure}[h]
		      	\centering
		      	\includegraphics[width=0.4\paperwidth]{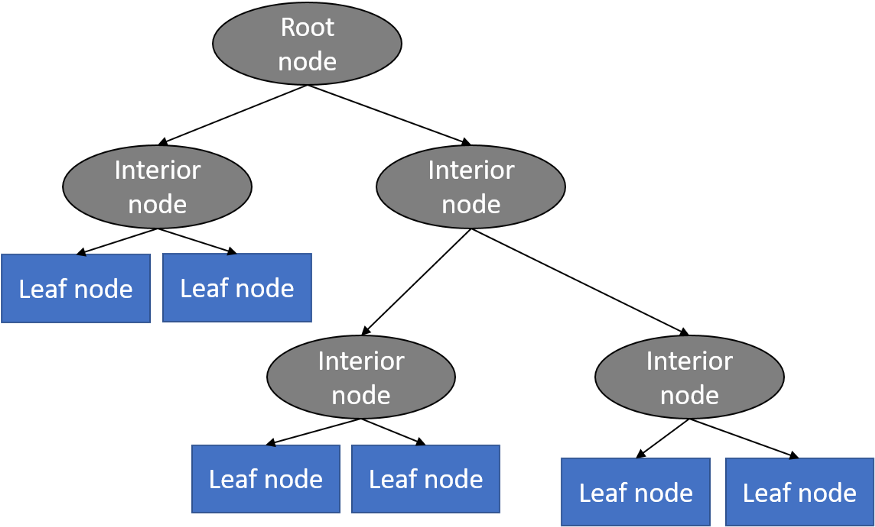}
		      	\caption{Architecture of ecision Tree Algorithm}
		      \end{figure}
		      \begin{enumerate}
		      	\item Decision tree \cite{dt} builds regression or classification models in the form of a tree structure. It breaks down a dataset into smaller and smaller subsets while at the same time an associated decision tree is incrementally developed. The result is a tree with decision nodes and leaf nodes.
		      	\item There are three distinct sorts of nodes in this regression tree. The Root Node is the primary node, representing the whole sample and potentially being subdivided into further nodes. Features of a dataset are represented by Interior Nodes, while decision rules are shown by Branches. In the end, the result is shown by the Leaf Nodes. If you have an issue that requires a choice, this algorithm is excellent. \cite{dt_website}
		      	\item A single data point is processed all the way down to the leaf node by asking and answering True/False queries. Ultimately, the dependent variable value in each leaf node is averaged to arrive at a final forecast. The Tree is able to provide an accurate prediction after going through several rounds.
		      	\item The benefits of using decision trees include their simplicity, the fact that they require less data cleaning, the fact that non-linearity has no effect on the performance of the model, and the fact that the number of hyper-parameters to be set is practically zero.
		      \end{enumerate}
		      		      
		\item \textit{K-Nearest Neighbors Regression}
		      \begin{figure}[h]
		      	\centering
		      	\includegraphics[width=0.3\paperwidth]{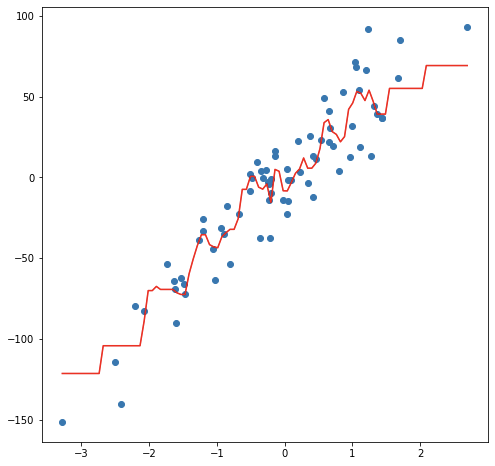}
		      	\caption{Example of KNN Algorithm}
		      \end{figure}
		      \begin{enumerate}
		      	\item K-Nearest Neighbors \cite{knn} is an easy-to-implement method that remembers past examples and makes a numerical prediction based on a similarity measure (e.g., distance functions). KNN is a non-parametric technique that has been utilised for statistical estimates and pattern recognition since the early 1970s.
		      	\item We need to use cross-validation to choose K. Unlike classification we cannot accuracy use as metric, since our predictions will almost never exactly match the true response variable values. Therefore in the context of KNN regression we will use root mean square prediction error (RMSPE) instead. The mathematical formula for calculating RMSPE is:
		      	      \begin{equation}
		      	      	RMSPE = \sqrt{\frac{\sum_{1}^{n}\left ( y_i - \hat{y_i} \right )^{2}}{n}}
		      	      \end{equation}
		      	      Where n is the number of observations, y\textsubscript{i} is the observed value for the i\textsuperscript{th} observation, and \^{y\textsubscript{i}} is the predicted value for the i\textsuperscript{th} observation.
		      	              	
		      	\item To put it another way, for each observation in our test (or validation) set, we calculate the squared difference between the anticipated and true response value, average over observations, and then square root. Since differences can be either positive or negative—that is, we can over- or under-estimate the genuine response value—we utilise the squared difference rather than merely the difference. \cite{knn_1}
		      	      		      
		      \end{enumerate}
	\end{enumerate}
		
	\section{Results}
	As we've seen, we have put our datasets through their paces with a wide range of ML algorithms. The Mean Absolute Error (MAE) represents the error that was introduced into our findings. The formula for the same is
		
	\begin{equation}
		MAE = \frac{\sum_{i=1}^{n} |y_i-x_i|)}{n} 
	\end{equation}

	\noindent Where, y\textsubscript{i} is the prediction, x\textsubscript{i} is the true value and n specifies the total number of data points.
	\newline
	\begin{table}
		\centering
		\caption{Mean Absolute Error Of Different Methods}\label{tab1}
		\begin{tabular}{|l|l|}
			\hline
			Method                        & Mean Absolute Error \\
			\hline
			Linear Regression             & 0.56              \\
			Random Forest Regressor       & 1.72              \\
			Decision Tree Regressor       & 1.49              \\
			K-Nearest Neighbour Regressor & 2.02              \\
			\hline
		\end{tabular}
	\end{table}
		
	\noindent It was observed that all 4 algorithms performed well considering how small the data set was. All algorithms gave promising results with Linear Regression lending the lowest MAE. Though with increasing data set, it would be wise to consider all the remaining models as well. The scores will only improve as the data set increases.

	\section{Conclusion}
	The results show us that the gross loss can be predicted to an error margin of ±0.5. The proof of concept needed to be derived from the results was sufficient to take to the company to act on it. Each of the 4 algorithms have potential, Linear Regression being the most promising one so far. Further testing needs to be done by increasing the number of data set and even expanding to different category of jewelry. The implementation of this innovation in the field of jewelry manufacturing would be a big undertaking and would be a time consuming and labour-intensive processes but one which would bare fruitful results.
		
	\bibliographystyle{unsrt}
	\bibliography{ref}
\end{document}